\documentclass[10pt, conference, compsocconf]{IEEEtran}
\usepackage{graphicx}
\usepackage{float}
\usepackage{multirow}

\ifCLASSOPTIONcompsoc
  \usepackage[nocompress]{cite}
\else
  \usepackage{cite}
\fi

\ifCLASSINFOpdf
 
\else
 
\fi

\usepackage{color}
\usepackage[table,xcdraw]{xcolor}

\usepackage{hyperref}
\hypersetup{
	bookmarks=false,
	colorlinks=true,       
	linkcolor=blue,          
	citecolor=green,        
	filecolor=magenta,      
	urlcolor=cyan         
}

\makeatletter
\newcommand{\newlineauthors}{%
  \end{@IEEEauthorhalign}\hfill\mbox{}\par
  \mbox{}\hfill\begin{@IEEEauthorhalign}
}
\makeatother

\newcounter{todocounter}

\usepackage{amsmath}
\usepackage{amssymb}

\title{A title}
\author{ \IEEEauthorblockN{Bao Hieu Tran \IEEEauthorrefmark{1}, Thanh Le-Cong  \IEEEauthorrefmark{1}, Huu Manh Nguyen, Duc Anh Le,  Thanh Hung Nguyen \IEEEauthorrefmark{2}, Phi Le Nguyen\IEEEauthorrefmark{2}}
    \IEEEauthorblockA{School of Information and Communication Technology}
    \IEEEauthorblockA{Hanoi University of Science and Technology} 
    \IEEEauthorblockA{Hanoi, Vietnam}
    \IEEEauthorblockA{\{hieu.tb167182, thanh.ld164834, manh.nh166428, anh.nd160126\}@sis.hust.edu.vn, \{lenp, hungnt\}@soict.hust.edu.vn}
}

\begin{document}

\title{SAFL: A Self-Attention Scene Text Recognizer with Focal Loss}

\maketitle
\begingroup\renewcommand\thefootnote{\IEEEauthorrefmark{1}}
\footnotetext{Authors contribute equally}
\begingroup\renewcommand\thefootnote{\IEEEauthorrefmark{2}}
\footnotetext{Corresponding author}
\begin{abstract}
In the last decades, scene text recognition has gained worldwide attention from both the academic community and actual users due to its importance in a wide range of applications.
Despite achievements in optical character recognition, scene text recognition remains challenging due to inherent problems such as distortions or irregular layout. Most of the existing approaches mainly leverage recurrence or convolution-based neural networks. However, while recurrent neural networks (RNNs) usually suffer from slow training speed due to sequential computation and encounter problems as vanishing gradient or bottleneck, CNN endures a trade-off between complexity and performance. 
In this paper, we introduce SAFL,  a self-attention-based neural network model with the focal loss for scene text recognition, to overcome the limitation of the existing approaches. 
The use of focal loss instead of negative log-likelihood helps the model focus more on low-frequency samples training.
Moreover, to deal with the distortions and irregular texts, we exploit Spatial TransformerNetwork (STN) to rectify text before passing to the recognition network.
We perform experiments to compare the performance of the proposed model with seven benchmarks. The numerical results show that our model achieves the best performance. 
\end{abstract}

\IEEEpeerreviewmaketitle

\section{Introduction}
In recent years, text recognition has attracted the attention of both academia and actual users due to its application on various domains such as translation in mixed reality, autonomous driving, or assistive technology for the blind. 
Text recognition can be classified into two main categories: scanned document recognition and scene text recognition. 
While the former has achieved significant advancements, the latter remains challenging due to scene texts' inherent characteristics such as the distortion and irregular shapes of the texts. 
Recent methods in scene text recognition are inspired by the success of deep learning-based recognition models. Generally, these methods can be classified in two approaches: recurrent neural networks (RNN) based and convolutional neural networks (CNN) based. RNN-based models have shown their effectiveness, thanks to capturing contextual information and dependencies between different patches. However, RNNs typically compute along with the symbol positions of the input and output sequences, which cannot be performed in parallel fashion, thus leads to high training time. Furthermore, RNNs also encounter problems such as vanishing gradient \cite{bengio1994learning} or bottleneck \cite{cho2014properties}. CNN-based approach, which allows computing the hidden representation parallelly, have been proposed to speed up the training procedure. However, to capture the dependencies between distant patches in long input sequences, CNN models require stacking more convolutional layers, which significantly increases the network's complexity. Therefore, CNN-based methods suffer the trade-off between complexity and accuracy. To remedy these limitations, in natural language processing (NLP) fields, a self-attention based mechanism named transformer \cite{vaswani2017attention} has been proposed. 
In the transformer, dependencies between different input and output positions are captured using a self-attention mechanism instead of sequential procedures in RNN. 
This mechanism allows more computation parallelization with higher performance. 
In the computer vision domain, some research have leveraged the transformer architecture and showed the effectiveness of some problems \cite{carion2020end} \cite{parmar2018image}
\begin{figure*}[bt]
    \centerline{\includegraphics[width=0.8\linewidth]{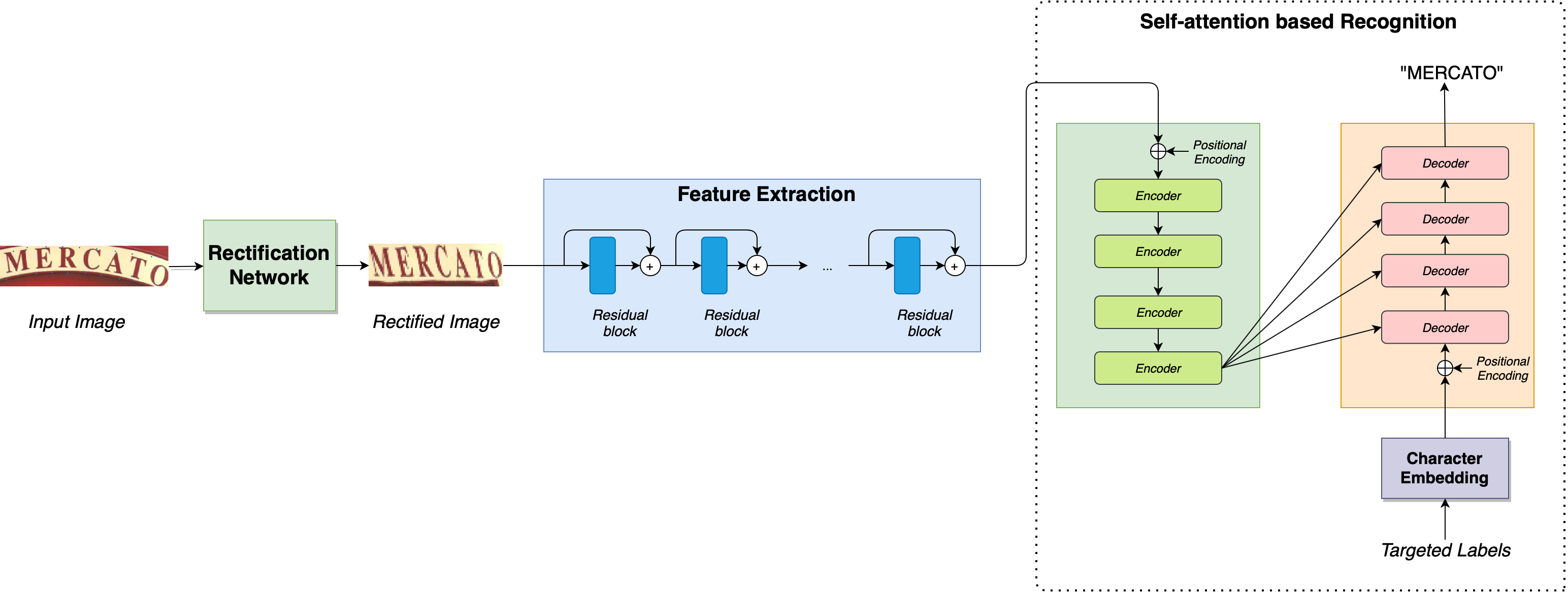}}
    \caption{Overview of SAFL}
    \label{fig:overall}
    \vspace{-5pt}
 \end{figure*}
 
Inspired by the transformer network, in this paper, we propose a self-attention based scene text recognizer with focal loss, namely as SAFL. Moreover, to tackle irregular shapes of scene texts, we also exploit a text rectification named Spatial Transformer Network (STN) to enhance the quality of text before passing to the recognition network. SAFL, as depicted in Figure \ref{fig:overall}, contains three components: rectification, feature extraction, and recognition. 
First, given an input image, the rectification network, built based on the Spatial Transformer Network (STN) \cite{jaderberg2015spatial}, transforms the image to rectify its text.
Then, the features of the rectified image are extracted using a convolutional neural network.
Finally, a self-attention based recognition network is applied to predict the output character sequence. Specifically, the recognition network is an encoder-decoder model, where the encoder utilizes multi-head self-attention to transform input sequence to hidden feature representation,
then the decoder applies another multi-head self-attention to output character sequence. 
To balance the training data for improving the prediction accuracy, we exploit focal loss instead of negative log-likelihood as in most recent works \cite{shi2016robust} \cite{shi2018aster}. 

To evaluate our proposed model's performance, we train SAFL with two synthetic datasets: Synth90k \cite{jaderberg2014synthetic} and SynthText \cite{gupta2016synthetic}, and compare its accuracy with standard benchmarks, on both regular and irregular datasets. The experiment results show that our method outperforms the state-of-the-art on all datasets. Furthermore, we also perform experiments to study the effectiveness of focal loss. The numerical results show the superiority of focal loss over the negative log-likelihood loss on all datasets. 

The remainder of the paper is organized as follows. Section \ref{sec:relatedwork} introduces related works. We describe the details of the proposed model in Section \ref{sec:proposed} and present the evaluation results in Section \ref{sec:experiment}. Finally, we conclude the paper and discuss the future works in Section \ref{sec:conclusion}.

\section{Related Work} \label{sec:relatedwork}
Scene text recognition has attracted great interest over the past few years. Comprehensive surveys for scene text recognition may be found in \cite{simonyan2014very} \cite{zhu2016scene} \cite{ye2014text}. As categorized by previous works \cite{shi2018aster} \cite{baek2019wrong} \cite{wang2019simple}, scene text may be divided into two categories: regular and irregular text. 
The regular text usually has a nearly horizontal shape, while the irregular text has an arbitrary shape, which may be distorted. 

\subsection{Regular text recognition} Early work mainly focused on regular text and used a bottom-up scheme, which first detects individual characters using a sliding window, then recognizing the characters using dynamic programming or lexicon search \cite{wang2011end} \cite{yao2014strokelets} \cite{wang2010word}. However, these methods have an inherent limitation, which is ignoring contextual dependencies between characters. Shi et al. \cite{shi2016end} and He et al. \cite{he2016reading} typically regard text recognition as a sequence-to-sequence problem. Input images and output texts are typically represented as patch sequences and character sequences, respectively. This technique allows leveraging deep learning techniques such as RNNs or CNNs to capture contextual dependencies between characters \cite{shi2016robust} \cite{shi2016end} \cite{he2016reading}, lead to significant improvements in accuracy on standard benchmarks. Therefore, recent work has shifted focus to the irregular text, a more challenging problem of scene text recognition.

\subsection{Irregular text recognition} Irregular text is a recent challenging problem of scene text recognition, which refers to texts with perspective distortions and arbitrary shape. 
The early works correct perspective distortions by using hand-craft features. However, these approaches require correct tunning by expert knowledge for achieving the best results because of a large variety of hyperparameters. Recently, Yang et al. \cite{yang2017learning} proposed an auxiliary dense character detection model and an alignment loss to effectively solve irregular text problems. Liu et al. \cite{liu2018char} introduced a Character-Aware Neural Network (Char-Net) to detect and rectify individual characters. Shi et al. \cite{shi2016robust} \cite{shi2018aster} addressed irregular text problems with a rectification network based on Spatial Transformer Network (STN), which transform input image for better recognition. Zhan et al. \cite{zhan2019esir} proposed a rectification network employing a novel line-fitting transformation and an iterative rectification pipeline for correction of perspective and curvature distortions of irregular texts. 
\section{Proposed model}\label{sec:proposed}
Figure \ref{fig:overall} shows the structure of SAFL, which is comprised of three main components: rectification, feature extraction, and recognition. 
The rectification module is a Spatial Transformer Network (STN) \cite{jaderberg2015spatial}, which receives the original image and rectifies the text to enhance the  quality. 
The feature extraction module is a convolution neural network that extracts the information of the rectified image and represents it into a vector sequence.
The final module, i.e., recognition, is based on the self-attention mechanism and the transformer network architecture \cite{vaswani2017attention}, to predict character sequence from the feature sequence. 
In the following, we first present the details of the three components in Section \ref{sec:transformation}, \ref{sec:feature} and \ref{sec:recognition}, respectively. Then, we describe the training strategy using focal loss in Section \ref{sec:loss}.

\subsection{Rectification} \label{sec:transformation}
In this module, we leverage a Thin Plate Spline (TPS) transformation \cite{shi2018aster}, a variant of STN, to construct a rectification network. 
Given the input image $I$ with an arbitrary size, the rectification module first resizes $I$ into a predefined fixed size.  
Then the module detects several control points along the top and bottom of the text's bounding. 
Finally, TPS applies a smooth spline interpolation between a set of control points to rectify the predicted region to obtain a fixed-size image.

\subsection{Feature Extraction} \label{sec:feature}
We exploit the convolution neural network (CNN) to extract the features of the rectified image (obtained from a rectification network) into a sequence of vectors. Specifically, the input image is passed through convolution layers (ConvNet) to produce a feature map. Then, the model separates the feature map by rows. The output received after separating the feature map are feature vectors arranged in sequences. The scene text recognition problem then becomes a sequence-to-sequence problem whose input is a sequence of characteristic vectors, and whose output is a sequence of characters predicted. Based on the proposal in \cite{vaswani2017attention}, we further improve information about the position of the text in the input image by using positional encoding. Each position $pos$ is represented by a  vector whose value of the $i^{th}$ dimension, i.e., ${PE}_{(pos,i)}$, is defined as 
\begin{equation}
    {PE}_{(pos,i)}=\left\{\begin{array}{l}
 \sin{\frac{pos}{10000^{\frac{2i}{d_{model}}}}}, ~ \text{if}  ~ 0 \leq i \leq \frac{d_{model}}{2} \\ 
 \cos{\frac{pos}{10000^{\frac{2i}{d_{model}}}}}, ~ \text{if}  ~ \frac{d_{model}}{2} \leq i \leq d_{model},
\end{array}\right.
\end{equation}
where $d_{model}$ is the vector size. 
The position information is added into the encoding vectors. 
\subsection{Self-attention based recognition network} \label{sec:recognition}
The architecture of the recognition network follows the encoder-decoder model. Both encoder blocks and decoder blocks are built based on the self-attention mechanism. We will briefly review this mechanism before describing each network's details.

\subsubsection{Self-attention mechanism} 
Self-attention is a mechanism that extracts the correlation between different positions of a single sequence to compute a representation of the sequence. In this paper, we utilize the scaled dot-product attention proposed in \cite{vaswani2017attention}. This mechanism consists of queries and keys of dimension $d_{k}$, and values of dimension $d_{v}$. Each query performs the dot product of all keys to obtain their correlation. Then, we obtain the weights on the values by using the softmax function. In practice, the keys, values, and queries are also packed together into matrices $ K $, $ V $ and $ Q $. The matrix of the outputs is computed as follow:
\begin{equation}
\emph{Attention}(Q, K, V)=\emph{softmax}\left(\frac{Q K^{T}}{\sqrt{d_{k}}}\right) V
\end{equation}
The dot product is scaled by $\frac{1}{\sqrt{d_{k}}}$ to alleviate the small softmax values which lead to extremely small gradients with large values of $d_{k}$. \cite{vaswani2017attention}.

\subsubsection{Encoder}
Encoder is a stack of $N_{e}$ blocks. Each block consists of two main layers. The first layer is a  multi-head attention layer, and the second layer is a fully-connected feed-forward layer. 
The multi-head attention layer is the combination of multiple outputs of the scale dot product attention. 
Each scale-dot product attention returns a matrix representing the feature sequences, which is called head attention. 
The combination of multiple head attentions to the multi-head attention allows our model to learn more representations of feature sequences, thereby increasing the diversity of the extracted information, and thereby enhance the performance. Multi-head attention can be formulated as follows:
\begin{equation}
{MultiHead}(Q, K, V)= {Concat}\left({head}_{1}, \ldots, {head}_{h}\right) W^{O}
\end{equation}
where ${head}_{i}={Attention}\left(Q W_{i}^{Q}, K W_{i}^{K}, V W_{i}^{V}\right)$, $h$ is the number of heads, $W_{i}^{Q} \in \mathbb{R}^{d_{\text {model }} \times d_{k}}, W_{i}^{K} \in \mathbb{R}^{d_{\text {model }} \times d_{k}}, W_{i}^{V} \in \mathbb{R}^{d_{\text {model }} \times d_{v}}$, $W^{O} \in \mathbb{R}^{hd_{v} \times d_{\text {model }}}$ are weight matrices. $d_{k}, d_{v}$ and $d_{model}$ are set to the same value.
Layer normalization \cite{ba2016layer} and residual connection \cite{he2016deep} are added into each main layer (i.e., multi-head attention layer and fully-connected layer) to improve the training effect. Specifically, the residual connections helps to decrease the loss of information in the backpropagation process, while the normalization makes the training process more stable. 
Consequently, the output of each main layer with the input $x$ can be represented as ${LayerNorm}(x+\text {Layer}(x))$, where $Layer(x)$ is the function implemented by the layer itself, and $LayerNorm()$ represents the normalization operation. The blocks of the encoder are stacked sequentially, i.e., the output of the previous block is the input of the following block. 

\subsubsection{Decoder}
The decoding process predicts the words in a sentence from left to right, starting with the $\langle start \rangle$ tag until encountering the $\langle end \rangle$ tag. The decoder is comprised of $N_{d}$ decoder blocks. Each block is also built based on multi-head attention and a fully connected layer. The multi-head attention in the decoder does not consider words that have not been predicted by weighting these positions with $-\infty$. Furthermore, the decoder uses additional multi-head attention that receives keys and values from the encoder and queries from the decoder. Finally, the decoder's output is converted into a probability distribution through a linear transformation and softmax function.

\subsection{Training} \label{sec:loss}
\begin{figure}[bt]
    \centerline{\includegraphics[width=1\linewidth]{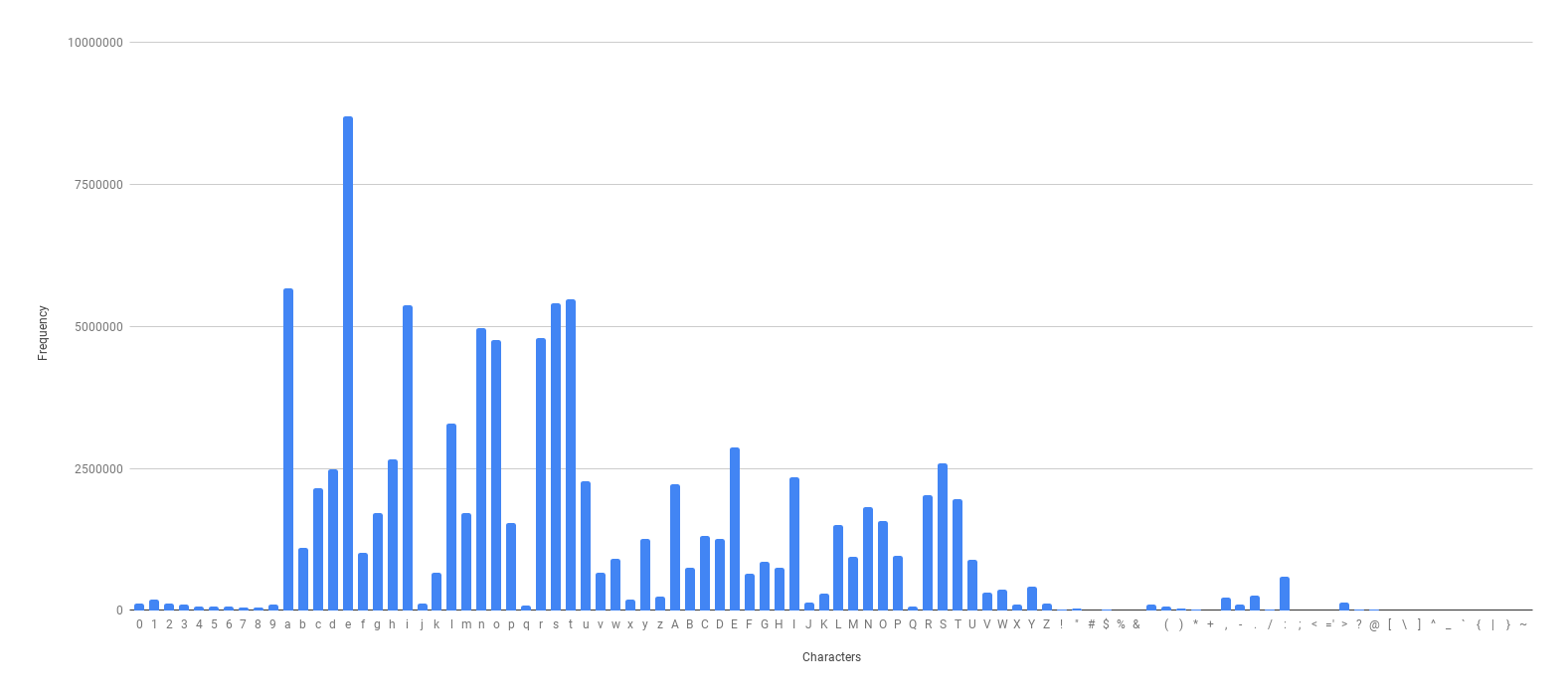}}
    \caption{Frenquency of characters in training lexicon}
    \label{fig:unbalance}
 \end{figure}
Figure \ref{fig:unbalance} shows that the lexicon of training datasets suffers from an unbalanced sample distribution. The unbalance may lead to severe overfitting for high-frequency samples and underfitting for low-frequency samples. To this end, we propose to use focal loss \cite{lin2017focal} instead of negative log-likelihood as in most of recent methods \cite{shi2016robust} \cite{shi2018aster}. By exploiting focal loss, the model will not encounter the phenomenon of ignoring to train low-frequency samples.

Focal loss is known as an effective loss function to address the unbalance of datasets. By reshaping the standard cross-entropy loss, focal loss reduces the impacts of high-frequency samples and thus focus training on low-frequency ones \cite{lin2017focal}. The focal loss is defined as follows:
\begin{equation}
F L\left(p_{t}\right)=-\alpha_{t}\left(1-p_{t}\right)^{\gamma} \log \left(p_{t}\right),
\end{equation}
where, $p_{t}$ is the probability of the predicted value, computed using softmax function, $\alpha$ and $\gamma$ are tunable hyperparameters used to balance the loss. 
Intuitively, focal loss is obtained by multiplying cross entropy by $\alpha_{t}\left(1-p_{t}\right)^{\gamma}$. 
Note that the weight $\alpha_{t}\left(1-p_{t}\right)^{\gamma}$ is inversely proportional with $p_{t}$, thus the focal loss helps to reduce the impact of high-frequency samples (whose value of $p_{t}$ is usually high) and focus more on low-frequency ones (which usually have low value of $p_{t}$).

Based on focal loss, we define our training objective as follows:
\begin{equation}
L=-\sum_{t=1}^{T}\left(\alpha_{t}\left(1- p\left(y_{t} \mid I\right))^{\gamma} \log p\left(y_{t} \mid I\right) \right))\right)
\end{equation}
where $y_{t}$ are the predicted characters, $T$ is the length of the predicted sequence, and $I$ is the input image. 
\section{Performance evaluation} \label{sec:experiment}
In this section, we conduct experiments to demonstrate the effectiveness of our proposed model. We first briefly introduce datasets used for training and testing, then we describe our implementation details. Next, we analyze the effect of focal loss on our model. Finally, we compare our model against state-of-the-art techniques on seven public benchmark datasets, including regular and irregular text. 
\subsection{Datasets} \label{sec:datasets}
The training datasets contains two datasets: \textbf{Synth90k} and \textbf{SynthText}. Synth90k is a synthetic dataset introduced in \cite{jaderberg2014synthetic}. This dataset contains 9 million images created by combining  90.000 common English words and random variations and effects. SynthText is a synthetic dataset introduced in \cite{gupta2016synthetic}, which contains 7 million samples by the same generation process as Synth90k \cite{jaderberg2014synthetic}. 
However, SynthText is targeted for text detection so that an image may contain several words. 
All experiments are evaluated on seven well-known public benchmarks described, which can be divided into two categories: regular text and irregular text. 
Regular text datasets include IIIT5K, SVT, ICDAR03, ICDAR13.
\begin{itemize}
    \item IIIT5K \cite{mishra2012top} contains 3000 test images collected from Google image searches. 
    \item ICDAR03 \cite{lucas2005icdar} contains 860 word-box cropped images.
    \item ICDAR13 \cite{karatzas2013icdar} contains 1015 word-box cropped images.
    \item SVT contains 647 testing word-box collected from Google Street View. 
\end{itemize}
{Irregular text} datasets include ICDAR15, SVT-P, CUTE.
\begin{itemize}
    \item ICDAR15 \cite{karatzas2015icdar} contains 1811 testing word-box cropped images collected from Google Glass without careful positioning and focusing. 
    \item SVT-P \cite{quy2013recognizing} contains 645 testing word-box cropped images collected from Google Street View. Most of them are heavily distorted by the non-frontal view angle. 
    \item CUTE \cite{risnumawan2014robust} contains 288 word-box cropped images, which are curved text images.
\end{itemize}
\subsection{Configurations}
\subsubsection{Implementation Detail}
We implement the proposed model by Pytorch library and Python programming language. The model is trained and tested on an NDIVIA RTX 2080 Ti GPU with $12$ GB memory. We train the model from scratch using Adam optimizer with the learning rate of $0.00002$. To evaluate the trained model, we use dataset III5K. The pretrained model and code are available at \cite{sourcecode}

\subsubsection{Rectification Network} 
All input images are resized to $64 \times 256$ before applying the rectification network. The rectification network consists of three components: a localization network, a thin plate spline (TPS) transformation,  and a sampler. The localization network consists of 6 convolutional layers with the kernel size of $3 \times 3$ and two fully-connected (FCN) layers. Each FCN is followed by a $2 \times 2$ max-pooling layer. The number of the output filters is 32, 64, 128, 256, and 256. The number of output units of FCN is 512 and 2K, respectively, where $K$ is the number of the control points. In all experiments, we set $K$ to 20, as suggested by \cite{shi2018aster}. The sampler generates the rectified image with a size of $32 \times 100$. The size of the rectified image is also the input size of the feature extraction module. 

\subsubsection{Feature Extraction} We construct the feature extraction module based on Resnet architecture \cite{he2016deep}. The configurations of the feature extraction network are listed in Table \ref{tab:FEarchitecture}. Our feature extraction network contains five blocks of 45 residual layers. Each residual unit consists of a $1 \times 1$ convolutional layer, followed by a $3 \times 3$ convolution layer. In the first two blocks, we use $2 \times 2$ stride to reduce the feature map dimension. In the next blocks, we use $2 \times 1$ stride to downsampled feature maps. The $2 \times 1$ stride also allows us to retain more information horizontally to distinguish neighbor characters effectively.

\begin{table}[h]
\centering
\caption{Feature extraction network configurations.Each block is a residual network block. "s" stands for stride of the first convolutional layer in a block.}
\label{tab:FEarchitecture}
\resizebox{.5 \textwidth}{!}{
\begin{tabular}{|c|c|c|c|}
\hline
 & Layer & Feature map size & Configuration \\ \hline
\multirow{6}{*}{Encoder} & 
Block 0 & $32 \times 100$ & $3 \times 3$ conv, s($1 \times 1$)  \\ \cline{2-4} 
& Block 1 & $16 \times 50$ & $\left[\begin{array}{l}1 \times 1 \operatorname{conv}, 32 \\ 3 \times 3 \operatorname{conv}, 32\end{array}\right] \times 3$, s($2 \times 2$)  \\ \cline{2-4} 
& Block 2 & $8 \times 25$ & $\left[\begin{array}{l}1 \times 1 \operatorname{conv}, 64 \\ 3 \times 3 \operatorname{conv}, 32\end{array}\right] \times 3$, s($2 \times 2$)   \\ \cline{2-4} 
& Block 3 & $4 \times 25$ & $\left[\begin{array}{l}1 \times 1 \operatorname{conv}, 128 \\ 3 \times 3 \operatorname{conv}, 32\end{array}\right] \times 3$, s($2 \times 1$)   \\ \cline{2-4} 
& Block 4 & $2 \times 25$ & $\left[\begin{array}{l}1 \times 1 \operatorname{conv}, 256 \\ 3 \times 3 \operatorname{conv}, 32\end{array}\right] \times 3$, s($2 \times 1$)   \\ \cline{2-4}
& Block 5 & $1 \times 25$ & $\left[\begin{array}{l}1 \times 1 \operatorname{conv}, 512 \\ 3 \times 3 \operatorname{conv}, 32\end{array}\right] \times 3$, s($2 \times 1$)   \\ \hline
\end{tabular}
}
\end{table}

\subsubsection{Recognition}
The number of blocks in the encoder and the decoder are set both to $4$.
In each block of the encoder and the decoder, the dimension of the feed forward vector and the ouput vector are set to $2048$ and $512$, respectively.
The number of head attention layers is set to $8$.
The decoder recognizes 94 different characters, including numbers, alphabet characters, uppercase, lowercase, and 32 punctuation in ASCII.

\subsection{Result and Discussion}
\subsubsection{Impact of focal loss}
To analyze the effect of focal loss, we study two variants of the proposed model. The first variant uses negative log-likelihood, and the second one leverages focal loss. 
\begin{table}[h]
\centering
\caption{Recognition accuracies with Negative log-likelihood and Focal Loss}
\label{tab:loss}
\begin{tabular}{|l|cc|}
\hline
\textbf{Variant} & \textbf{Negative log-likelihood} & \textbf{Focal Loss} \\ \hline
IIIT5K  & 92.6     & \textbf{93.9}\\
SVT  & 85.8     & \textbf{88.6}\\
ICDAR03 & 94.1     & \textbf{95}\\
ICDAR13 & 92& \textbf{92.8}\\
ICDAR15 & 76.1     & \textbf{77.5}\\
SVT-P& 79.4     & \textbf{81.7}\\
CUTE & 80.6     & \textbf{85.4}\\ \hline
Avarage & 86.9 & \textbf{88.2}\\ \hline

\end{tabular}
\end{table}
\par As shown in Table \ref{tab:loss}, the model with focal loss outperforms the one with log-likelihood on all datasets. Notably, on average, focal loss improves the accuracy by 2.3 \% compared to log-likelihood. For the best case, i.e., CUTE, the performance gap between the two variants is 4.8 \%
\subsubsection{Impact of rectification network}
In this section, we study the effect of text rectification by comparing SAFL and a variant which does not include the rectification module. 
\begin{table}[h]
\centering
\caption{Recognition accuracies with and without rectification}
\label{tab:withoutrec}
\begin{tabular}{|l|cc|}
\hline
\textbf{Variant} & \textbf{SAFL w/o text rectification} & \textbf{SAFL} \\ \hline
IIIT5K  & 90.7   & \textbf{93.9}\\
SVT  & 83.3    & \textbf{88.6}\\
ICDAR03 & 93     & \textbf{95}\\
ICDAR13 & 90.7& \textbf{92.8}\\
ICDAR15 & 72.9     & \textbf{77.5}\\
SVT-P& 71.6    & \textbf{81.7}\\
CUTE & 77.4     & \textbf{85.4}\\ \hline
Avarage & 84.1 & \textbf{88.2}\\ \hline

\end{tabular}
\end{table}

Table \ref{tab:withoutrec} depicts the recognition accuracies of the two models over seven datasets. It can be observed that the rectification module increases the accuracy significantly. Specifically, the performance gap between SAFL and the one without the rectification module is $4.1 \%$ on average. 
In the best cases, SAFL improves the accuracy by $10.1\%$ and $7\%$ compared to the other on the datasets SVT-P and CUTE, respectively.
The reason is that both SVT-P and CUTE contains many both irregular texts such as perspective texts or curved texts. 

\subsubsection{Comparison with State-of-the-art}
In this section, we compare the performance of SAFL with the latest approaches in scene text recognition. 
The evaluation results are shown in Table \ref{tab:CompareSOTA}.
In each column, the best value is bolded. 
the "Avarage" column is the weighted average over all the data sets. 
Concerning the irregular text, it can be observed that SAFL achieves the best performance on $3$ data sets. 
Particularly, SAFL outperforms the current state-of-the-art, ESIR \cite{zhan2019esir}, by a margin of $1.2 \%$ on average, particulary on CUTE ($+2.1\%$) and SVT-P ($+2.1\%$).
Concerning the regular datasets, SAFL outperforms the other methods on two datasets IIIT5K and ICDAR03. Moreover, SAFL also shows the highest average accuracy over all the regular text datasets. To summarize, SAFL achieves the best performance on 5 of 7 datasets and the highest average accuracy on both irregular and regular texts. 
\begin{table*}[t]
  \centering
\caption{Scene text accurancies (\%) over seven public benchmark test datasets.}
\label{tab:CompareSOTA}
  \begin{tabular}{|l|cccc||c|ccc||c|}
\hline
\multicolumn{1}{|c|}{\multirow{2}{*}{\textbf{Method}}} & \multicolumn{5}{c|}{\textbf{Regular test dataset}}& \multicolumn{4}{c|}{\textbf{Irregular test dataset}} \\ \cline{2-10} 
\multicolumn{1}{|c|}{} & IIIT5k & SVT & ICDAR03 & ICDAR13 &Average & ICDAR15 & SVT-P & CUTE & Average  \\ \hline
Jaderberg et al. \cite{jaderberg2014deep} & - & 80.7 & 93.1  & 90.8  & - & - & - & - & - \\
CRNN \cite{shi2016end} & 78.2 & 80.8 & 89.4 & 86.7 & 81.8 & - & - & - & - \\ 
RARE \cite{shi2016robust} & 81.9 & 81.9 & 90.1 & 88.6 & 85.3 & - & 71.8 & 59.2 & - \\ 
Lee et al. & 78.4 & 80.7& 88.7 & 90.0 & 82.4 & - & - & - & - \\ 
Yang et al. \cite{yang2017learning} & - & 75.8 & - & - & - & - & 75.8 & 69.3 & - \\ 
FAN \cite{cheng2017focusing} & 87.4 & 85.9 & 94.2 & 93.3 & 89.4 & 70.6 & - & - & - \\
Shi et al. \cite{shi2016robust} & 81.2 & 82.7 & 91.9 & 89.6 & 84.6 & - & - & - & - \\
Yang et al. \cite{yang2017learning} & - & - & - & - & - & - & 75.8 & 69.3 & - \\ 
Char-Net \cite{liu2018char}  & 83.6 & 84.4 & 91.5 & 90.8 & 86.2 & 60.0 & 73.5 & - & - \\ 
AON \cite{cheng2018aon}& 87.0 & 82.8 & 91.5 & - & - & 68.2 & 73.0 & 76.8 & 70.0 \\
EP \cite{bai2018edit}& 88.3 & 87.5 & 94.6 &\textbf{ 94.4} & 90.3 & 73.9 & - & - & - \\
Liao et al. \cite{liao2019scene} & 91.9 & 86.4 & - & 86.4 & - & - & - & 79.9 & - \\
Baek et al. \cite{baek2019wrong} & 87.9 & 87.5 & 94.9 & 92.3 & 89.8 & 71.8 & 79.2 & 74.0 & 73.6 \\ 
ASTER \cite{shi2018aster}& 93.4 & 89.5 & 94.5 & 91.8 & 92.8 & 76.1 & 78.5 & 79.5 & 76.9 \\ 
SAR \cite{li2019show}& 91.5 & 84.5 & - & 91.0 & - & 69.2 & 76.4 & 83.3 & 72.1 \\
ESIR \cite{zhan2019esir} & 93.3 & \textbf{90.2} & - & 91.3 & - & 76.9 & 79.6 & 83.3 & 78.1 \\
\hline
\textbf{SAFL}   & \textbf{93.9} & 88.6 & \textbf{95}& 92.8 & \textbf{93.3} & \textbf{77.5} & \textbf{81.7} & \textbf{85.4} & \textbf{79.3} \\ \hline
\end{tabular}
\end{table*}

\section{Conclusion} \label{sec:conclusion}
In this paper, we proposed SAFL, a deep learning model for scene text recognition, which exploits self-attention mechanism and focal loss. 
The experiment results showed that SAFL achieves the highest average accuracy on both the regular datasets and irregular datasets. 
Moreover, SAFL outperforms the state-of-the-art on CUTE dataset by a margin of $2.1\%$. Summary, SAFL shows superior performance on 5 out of 7 benchmarks, including IIIT5k, ICDAR 2003, ICDAR 2015, SVT-P and CUTE.

\ifCLASSOPTIONcompsoc
  \section*{Acknowledgments}
\else
  \section*{Acknowledgment}
\fi
We would like to thank AIMENEXT Co., Ltd. for supporting our
research.

\bibliographystyle{IEEEtran}
\bibliography{references.bib}

\end{document}